\documentclass{article}

\usepackage[final,nonatbib]{apt_2021}

\usepackage{paralist}
\usepackage[utf8]{inputenc} 
\usepackage[T1]{fontenc}    
\usepackage{hyperref}       
\usepackage{url,multirow}            
\usepackage{booktabs}       
\usepackage{amsfonts}       
\usepackage{nicefrac}       
\usepackage{microtype}      
\usepackage{graphicx}
\usepackage{caption}
\usepackage{subcaption}
\graphicspath{{images/}}    

\title{Miutsu: NTU's TaskBot for the Alexa Prize}

\author{
	Yen-Ting Lin$^\star$\quad Hui-Chi Kuo\quad Ze-Song Xu\quad Ssu Chiu\\
	\bf Chieh-Chi Hung\quad Yi-Cheng Chen\quad Chao-Wei Huang\quad Yun-Nung Chen$^\dag$\\
	National Taiwan University, Taipei, Taiwan\\
	\texttt{$^\star$f08944064@csie.ntu.edu.tw\quad $^\dag$y.v.chen@ieee.org}
}

\begin{document}

\maketitle

\begin{abstract}
This paper introduces Miutsu, National Taiwan University’s Alexa Prize TaskBot, which is designed to assist users in completing tasks requiring multiple steps and decisions in two different domains -- home improvement and cooking. We overview our system design and architectural goals, and detail the proposed core elements, including question answering, task retrieval, social chatting, and various conversational modules. A dialogue flow is proposed to provide a robust and engaging conversation when handling complex tasks. We discuss the faced challenges during the competition and potential future work.
\end{abstract}

\section{Introduction}\label{sec:introduction}
Recent advances in artificial intelligence have led to the development of virtual assistants and revolutionized the way people interact with their digital devices. These assistants are capable of understanding natural language and are increasingly becoming more capable of completing various tasks, including travel planning, ordering groceries, providing weather information, etc. However, completing complex tasks is still challenging, because complex tasks may require multiple steps and decisions, such as finding a recipe, cooking food, or planning a home improvement project.

Therefore, this paper focuses on developing a virtual assistant capable of assisting users in completing complex tasks in two domains - \emph{home improvement} and \emph{cooking}. Our proposed virtual assistant is called ``Miutsu'' in the Alexa Prize competition, which is inspired from ``Mewtwo'', a fictional creature in Pokemon.
In the home improvement domain, Miutsu can help users find the right tools and products and provide step-by-step instructions on how to complete a home improvement project. Miutsu can help users find recipes, prepare food, and cook food in the cooking domain. Miutsu is designed to help users in both domains by providing information, answering questions, and recommendations.

\section{System Overview}\label{sec:system-overview}

\begin{figure}[t!]
  \centering
  \includegraphics[width=.85\linewidth]{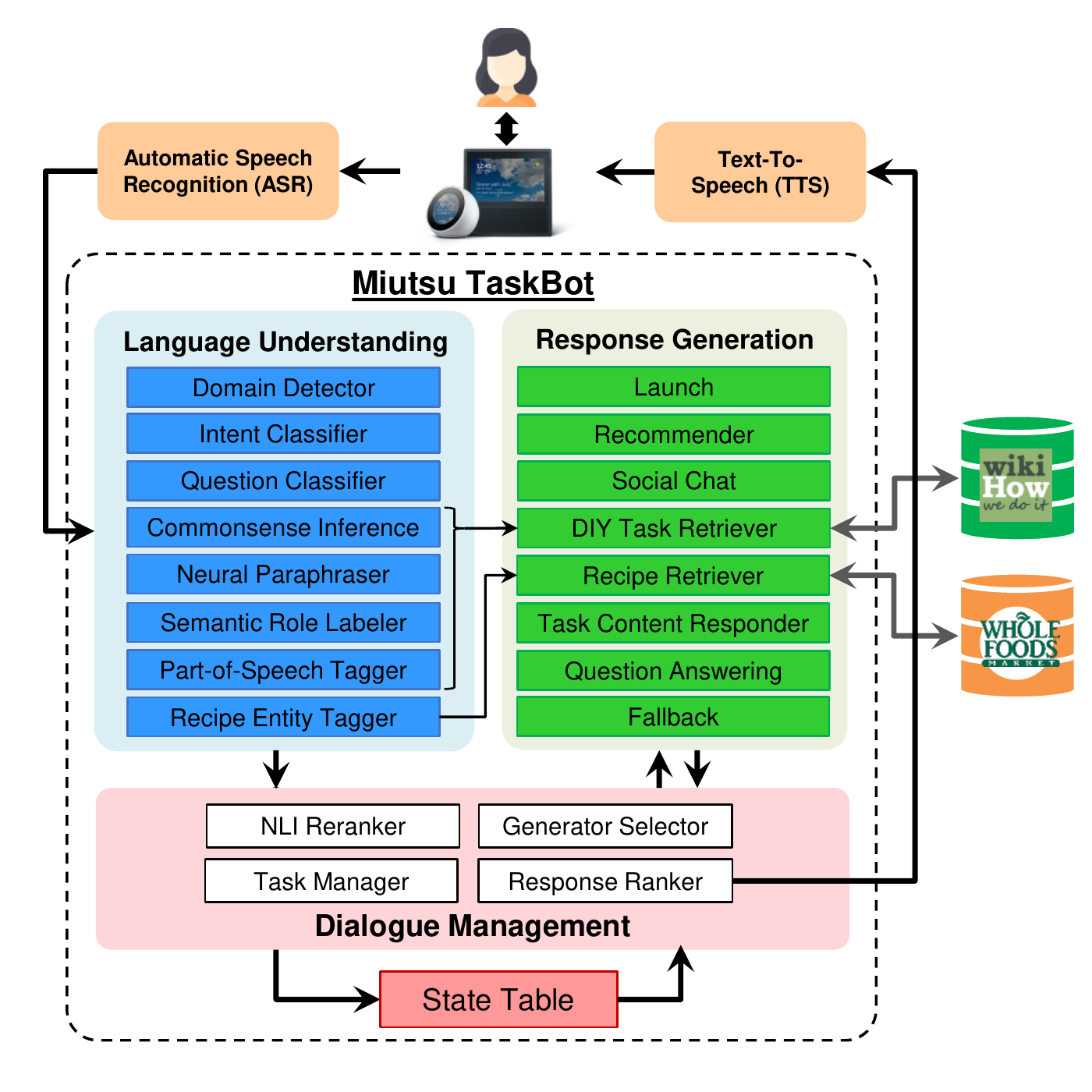}
  \vspace{-5mm}
  \caption{Illustration of our system architecture.}
  \label{fig:system_design}
  \vspace{-5mm}
\end{figure}

Figure \ref{fig:system_design} illustrates our bot framework, which contains three parts: 1) a language understanding component, 2) a dialogue management component, and 3) a response generation component. The user's spoken utterance is first transcribed into text via an Alexa Skills Kit (ASK) supplied automatic speech recognition (ASR), and the transcribed text is then passed to our framework to generate text responses in the Speech Synthesis Markup Language (SSML) format. The response is returned to the user via ASK's Text-To-Speech (TTS). The procedure of each component is detailed below.

The \textbf{language understanding} component (see Section \ref{sec:nlp-pipeline}) is to produce linguistic, semantic, and dialogue-related labels based on the user input and the current state. All modules in this component are deep learning models (except the PoS tagger), are hosted on EC2 GPU instances. PoS tagging is hosted on the Lambda function. Modules are stacked as pipelines, allowing modules to use annotations from others. The pipelines are run in parallel, and the results are combined to produce final labels for the next component.

Next, the \textbf{dialogue management} (see Section \ref{sec:dialogue-flow}) phase controls the conversation flow and decides which task to execute and how. The dialogue decision is based on various control modules: a generator selector, a response ranker, a task manager, and an NLI reranker. 
The generator selector runs a collection of \textbf{response generators} (see Section \ref{sec:response-generator}), which handle different tasks such as generating suggestions, chatting, presenting candidate tasks, providing task content, and answering questions. The response ranker chooses the best response among those produced by the selected response generators and sends the response to the user. The task manager controls the task flow and decides which task to run, how to navigate through the task flow, and when to terminate the task. The NLI reranker uses the user's utterance to rerank the candidate tasks during task retrieval.

Our system is built on top of the CoBot framework \cite{khatri2018advancing}, and all of our modules are stateless. These modules do not maintain any dialogue state internally. Instead, they use a DynamoDB state table to store and retrieve annotations, responses, and other dialogues.

\section{Dialogue Flow}\label{sec:dialogue-flow}
This section describes how our approach helps users find, select, and complete a home improvement task or follow a recipe. A three-phase process is implemented for a dialogue, where the user first starts the dialogue with a goal, then selects a DIY task or a recipe, and finally completes it.

\subsection{Dialogue Initialization}
At the beginning of a dialogue, the user usually initiates the dialogue by indicating their goal, either explicitly (e.g., ``How to repair a window'') or implicitly (e.g., ``I need to make a cake for a party''). However, users may also initiate the dialogue by asking a general question, such as ``What can you do?'' or ``How can you help me?''. In order to support different user scenarios, we identify \textit{user initiative} as a significant source of variation in user utterances. To handle the variation, we use domain and intent predictions from language understanding modules to classify user utterances into following three types:
\begin{compactitem}
    \item \textbf{High-initiative} user utterances where the users explicitly or implicitly indicate their goal, e.g. ``How to plant a vegetable garden'' or ``I want to make lemon pie''. These utterances mention a specific task, recipe, or ingredients, enabling the associated downstream task or recipe retrieval (detailed in \ref{ssec:diy_understanding} and \ref{subsec:recipe-retriever}).
    \item \textbf{Low-initiative} user utterances where the users either ask a general question, chitchat with our bot or do not contain any meaningful content~\cite{paranjape2020neural}, such as ``What can you do?''. These utterances do not convey any specific goal or intent. In these cases, we respond to the users with social conversations about what we can do via a social chat responder (see \ref{subsec:neural-chat}) and then elicit user initiative again.
    \item \textbf{Recommendation request} user utterances where users usually indicate specific domains but no specific goal, like ``Could you recommend a DIY project?''. We then suggest a DIY task or a recipe to users (see \ref{subsec:recommender}).
\end{compactitem}

\subsection{DIY Task / Recipe Selection}
Once the user indicates their goal, we use the goal as a constraint to retrieve corresponding tasks or recipes. The user can select a task or a recipe from a list of retrieved candidates. The user can indicate their choice by \emph{item name}, \emph{item order}, or \emph{touching the screen directly}. To alleviate ASR errors, when identifying the task/recipe in the user utterance, we use Levenshtein distance to select the most similar one. We also confirm with the user when the distance is greater than a threshold.

\subsection{Task Completion}
After the user confirms the task or recipe selection, we will provide instructions to the user. The instructions are generated using a template-based method (see Section \ref{subsubsec:task-responder}). The user navigates the instructions using a list of commands or touch screens. 
The system detects the user's intent though language understanding and keyword-based methods and then provides the corresponding response.
Note that users can ask questions during the insteraction, the system performs question answering detailed in Section \ref{subsec:question-answering}. The dialogue ends either when the user completes the task or explicitly terminates the interaction.

\section{Language Understanding}\label{sec:nlp-pipeline}


\subsection{Domain Detector}\label{subsec:domain-classifier}
The domain detector is to detect which topic the user utterance belongs to.
The main purpose is to tell if the user wants to perform DIY or cooking tasks, and it also helps to detect domains inappropriate for our bot to chat about.

\vspace{-3mm}
\paragraph{Data} 
There are 8 domains covered by our system, including 1) \textit{DIY}, 2) \textit{cooking}--which are the target domains defined in the challenge, and 3) \textit{finance}, 4) \textit{medicine}, 5) \textit{law}, 6) \textit{harm} (including dangerous and self-harm topics), 7) \textit{pornography, gambling or drugs}, 8) \textit{love or relationship}--which are the domains our bot should avoid chatting about.

Due to lack of real user utterances, to train the model in a supervised way, we first crawl the suitable data from the web, where two types of user sentences are considered.
\begin{compactitem}
    \item Interrogative sentences: all questions from the corresponding topics in Quora\footnote{\url{https://www.quora.com/}} (e.g. the topics ``recipes'' and ``do it yourself'' are for \textit{cooking}, and \textit{DIY} respectively).
    \item Non-interrogative sentences: all article titles from the corresponding subreddits\footnote{\url{https://www.reddit.com/}} (e.g. ``r/cooking'' and ``r/DIY'' for \textit{cooking} and \textit{DIY} respectively).\footnote{We use Praw (\url{https://github.com/praw-dev/praw}) to crawl data from Reddit.}
    For the \textit{DIY} domain, we additionally crawl article titles in WikiHow to better fit the target task retriever.
\end{compactitem}
In addition to the above 8 domain, we collect data for ``\emph{out-of-domain}'' detection, which includes questions and article titles from other popular topics and subreddits such as ``Technology'' in Quora and ``r/AskReddit'' subreddit. 
There are total 45K questions/titles, and split them into 80\%/20\% for training/validation.
In order to further test the performance of our model,  we hand-craft 64 sentences simulating possible user utterances.

\vspace{-3mm}
\paragraph{NLI-Based Model}
Different from most prior approaches that used a multi-label classifier to decide the domain the input utterance belongs to, we leverage the capability of natural language inference to decide the domain.
The reason is that our collected data is not in a large scale, so leveraging pre-trained capability may achieve better performance of domain detection.
Specifically, we formulate the domain detection task as a natural language inference (NLI) problem: Each question or title is a \emph{premise}, and the \emph{hypothesis} is ``This text is about \textit{X}'', where \textit{X} is the one of the defined domains or \textit{others} for out-of-domain.
We decide the domain based on if the outputted label is ``entailment'' or ``contradiction''.
We use a BART \cite{lewis-etal-2020-bart} model pretrained on MNLI~\cite{williams-etal-2018-broad}, and fine-tune it with our collected training samples in the NLI format.
The fine-tuned model achieved 94\% accuracy (60 out of 64) on our testing set.

\vspace{-3mm}
\paragraph{Domain Classification Model}
We also try the classical multi-label classification for deciding the domain using our collected data.
To address the issue about data scarcity, we augment the training data by utilizing the pseudo domain labels of all questions in Quora\footnote{\url{https://huggingface.co/datasets/quora}} predicted by the fine-tuned BART mentioned above.
Here a BERT \cite{devlin-etal-2019-bert} classifier is adopted, and the model achieves a 84\% accuracy (54 out of 64).

Comparing between above models, we use the NLI-based one as our final domain detector.

\subsection{Intent Classifier}\label{subsec:intent-classifier}
The intent classifier is to predict the user's intention for better generating responses. For example, when the user says ``I'm looking for some recipes. Any suggestion?'', then the user is looking for some suggestions, and the bot should try to find some recipes to recommend. On the other hand, in the same \textit{cooking} domain, ``I want to make chocolate cake.'' implies that the user is making a specific request, so the bot should directly search for the recipes of chocolate cake.

In addition to the default intents pre-defined by Amazon, we additionally defined 3 intents: 1) \textit{recommend}--where the user asks for task recommendation, 2) \textit{request}--where the user asks for a certain task, and 3) \textit{jump steps}--where user wants to jump to a specific step when performing a task. 
We manually craft about 100 examples for the newly defined intents, and train a BERT model on it. If the user utterance does not belong to any Amazon pre-defined intents, then the bot would use the trained model to predict the intent.

\subsection{Question Detector}\label{subsec:question-classifier}

Users ask questions in various ways, and some questions might even be hard to detect as questions. Recognizing questions is an essential part of question answering and controlling dialogue flow.
To better capture if the user is asking something, we build a question detection model, a binary classification model fine-tuned on a dataset of question and statement utterances. The question and statement labels are from the MIDAS dataset \cite{yu-yu-2021-midas}, which is a dialogue action classification dataset released by the previous socialbot team Gunrock.

\begin{figure}[t!]
  \centering
  \includegraphics[width=.9\linewidth]{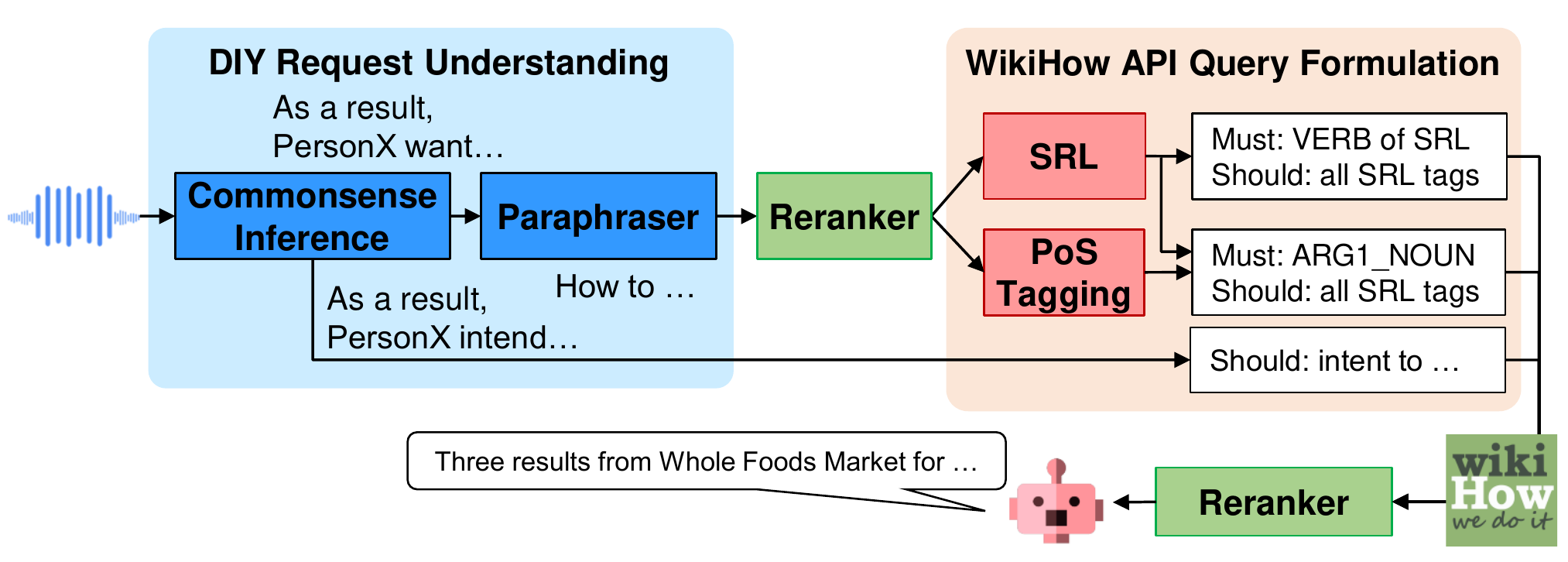}
  \vspace{-1mm}
  \caption{Illustration of the DIY task understanding framework.}
  \label{fig:DIY_Task_Retriever_Framework}
  \vspace{-2mm}
\end{figure}

\begin{figure}[t!]
  \centering
  \includegraphics[width=\linewidth]{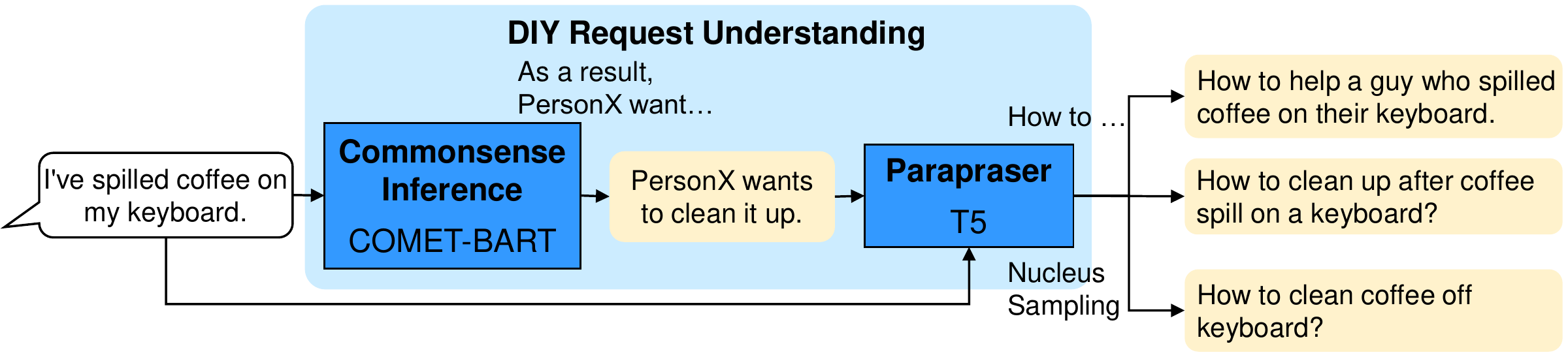}
  \vspace{-2.5mm}
  \caption{An example of the DIY request understanding module.}
  \label{fig:comet_xWant}
  \vspace{-3mm}
\end{figure}

\subsection{DIY Task Understanding}
\label{ssec:diy_understanding}

Figure~\ref{fig:DIY_Task_Retriever_Framework} illustrates the DIY task understanding framework, where two procedures are involved: request understanding and API query formulation.
In request understanding, we apply commonsense inference to infer the implicit user intent and then paraphrase it into the sentence better fitting the Wikihow backend.
To search the desired results with the fixed API functionality, we categorize DIY query formats into \emph{action-centric} and \emph{object-centric}.
Therefore, semantic role labeling and part-of-speech tagging is utilized to identify the important words for query formulation.


\subsubsection{Commonsense Inference}
\label{subsec:commonsense-inference}
Human chatting usually contain commonsense knowledge; specifically, human can understand implied semantics behind the sentences and then decide the responses.
For example, ``My roof is broken.'' implies that the user wants the roof to be fixed, but such inference requires human-level commonsense understanding, which is not easy to derive.
Even though a bot nowadays is capable of naturally talking to people, it usually tends to generate a general response without considering the implicit knowledge due to the complexity of human commonsense.
Inspired by this idea, we propose a commonsense inference module in our system, where COMET \cite{bosselut-etal-2019-comet} is employed and it memorizes both social and physical aspects of common human everyday experiences for inferring the potential intents.
We apply COMET-BART \cite{lewis-etal-2020-bart,bosselut-etal-2019-comet}, a seq2seq model trained on commonsense knowledge graphs, and two commonsense relations we use to infer the implicit user intents:
\begin{compactitem}
    \item ``xWant'' (As a result, PersonX wants): transforms a user utterance into what the user might truly desire when it is vague. An example is shown in the left part of Figure~\ref{fig:comet_xWant}.
    \item  ``xIntent'' (PersonX then): transforms better when incomprehensible and incomplete sentences due to ASR errors.
    Take an incomprehensible utterance from real logs ``how tall can nice my bedroom'' as an example, ``xWant'' outputs ``how to decorate'' and ``xIntent'' outputs ``to have a nice room''; we can easily see that the latter is more in line with the user utterance.
\end{compactitem}

\subsubsection{Neural Paraphraser}\label{subsec:neural-paraphraser}
Considering our backend database is better to handle the ``How to'' queries due to the Wikihow-supported API, we propose a neural-based sentence paraphraser to convert the input sentences into ones better fitting our API.
Specifically, we apply a T5 model fine-tuned on duplicated question pairs in QQP data, to convert a user query into a sentence starting with ``How to'' by fixing the first two words in decoding.
Nucleus sampling is used to generating the following tokens for better balancing diversity and fluency, where we sample 3 sentences from the paraphraser as shown in Figure~\ref{fig:comet_xWant}.
This procedure also helps the semantic role labeler easily find the focused part in the sentence.


Afterwards, to check if the paraphrased utterances are implied by the user utterance, we apply an NLI re-ranker to estimate $P(entailment)$ as the ranking score. The NLI model we used is \texttt{roberta-large-mnli}, where the \emph{premise} is a user utterance and the \emph{hypothesis} is the paraphrased utterance.
Hence, a higher ranking score implies a better paraphrased result.



\subsubsection{Semantic Role Labeling}\label{subsec:semantic-role-labeler}
Semantic role labeling do shallow semantic parse of natural language texts that produces predicate-argument structures of sentences.
The wikihow API-returned results are poor when directly inputting the ``How to'' sentences, because the API may not be able to figure out which part is more important for searching.
To address this problem, we apply semantic role labeling to extract the important parts by semantic roles such as \emph{verb} and its first argument.
Here a well-trained BERT-based model \cite{shi2019simple} is employed, which is a current state-of-the-art single model for English PropBank SRL (Newswire sentences) provided by AllenNLP\footnote{\url{https://allenai.org/allennlp}}.
It truly helps us to filter out irrelevant information.

However, some results may be still noisy if the argument contains an adjective, since the API focuses on the adjective instead of the noun of the argument.
To solve this problem, we further apply part-of-speech tagging on the paraphrased sentence for better handling such scenarios.
Here we use the same model as SRL to tokenize the sentence and obtain PoS tags of all words. Then we can extract the noun of ARG1 as our key object. 
As shown in the right part of Figure~\ref{fig:DIY_Task_Retriever_Framework}, the \emph{action-centric} query puts the \emph{verb} word as the \textbf{Must} constraint during search, while the \emph{object-centric} query puts \emph{noun} of ARG1 instead.
In addition to the key action or object, the rest words of the input sentence are put in the \textbf{Should} constraint of the API to capture minor information.

\subsubsection{Part-of-Speech (PoS) Tagging}\label{subsec:pos-tagger}
Part-of-speech tagging is to classify words into their associated PoS tags, which contain eight common tags: \emph{nouns}, \emph{verbs}, \emph{adjectives}, \emph{adverbs}, \emph{pronouns}, \emph{conjunctions}, \emph{prepositions} and \emph{interjections}.
The motivation behind using PoS tags is to decide which word contains the main concept so that we can formulate the API query accordingly.
In more details, the words annotated with \emph{noun} are treated as the core information for searching.
To compensate for the incompleteness of implicit intent utterances, the bottom line (the third query) takes the commonsense-inferred ``xIntent'' relation.

\subsection{Recipe Entity Tagging}\label{subsec:recipe-entity-tagger}
A recipe entity tagger aims at tagging recipe query information to search for recipes that suit user's preference.
The Whole Foods Market recipe retrieval API accepts several constraints--dish name, ingredients, cuisines, and occasion.
We design two types of taggers to identify these entities: a \emph{dish name} tagger specific for recipe names, and a general tagger for five cooking-related entity types: \textit{ingredients}, \textit{negative-ingredient}, \textit{cuisines}, \textit{meal course}, and \textit{occasion}.
In particular, \emph{negative-ingredient} represents the unwanted ingredient.
The reason of performing two taggers is that the dish name may overlap with other entities.
For example, ``I want to make a strawberry cupcake without chocolate,'' the dish name tagger tags \textit{strawberry cupcake}, and the general tagger tags \textit{strawberry} as an ingredient and \textit{chocolate} as a negative ingredient.

Our recipe entity tagger is a BERT-based model fine-tuned on our manually annotated data in the IOB format.
We manually craft 30 utterances as templates of users requesting recipes, and then crawl and annotate entities in titles from Quora spaces related to food\footnote{\url{https://www.quora.com/topic/Cooking}}.
We bootstrap the dataset by infilling the entities present in the crawled dataset into manual templates.
Our dish name tagger achieves 86\% in F1, and the cooking entity tagger achieves 59\% in F1 evaluated in our collected testing split.

\section{Response Generation}\label{sec:response-generator}

This section describes our approach to generate responses given language understanding annotations and the dialogue state. The response generator is referred as RG below.

\subsection{Launch and Recommender}\label{sec:launch} \label{subsec:recommender}
The launch RG handles the first turn of a dialogue, where it starts to greet the user and proactively suggest a task or a recipe. Following the first turn, if the user chooses to request a recommendation, the recommender RG then provides a list of recommended tasks or recipes.
The user can then select one from the list. 
To produce a natural language prompt for each suggested task, we paraphrase the task name into a conversational-styled sentence using GPT-3 \cite{brown2020language}\footnote{\textit{GPT3: text-davinci-001}} with a template prompt\footnote{``Write an interesting, engaging and conversational sentence for the DIY task: \textit{TASK\_NAME}.''}.

\subsection{Social Chat}\label{subsec:neural-chat}
To handle user utterances unrelated to our two target domains at the first few turns, a social char RG is proposed to generate social conversation responses to users.
Here we 
In order to use social chatting and simultaneously guide the dialogue to our target DIY and cooking domains, we embed the DIY and cooking related persona profile (the sentence ``\textit{I enjoy helping people with cooking and home improvement projects.}'' is prepended in the conversation context) into BlenderBot \cite{roller2020recipes}\footnote{\url{https://huggingface.co/facebook/blenderbot-400M-distill}} for generating suitable social responses.

Our neural generative model has several frequent drawbacks which negatively impact the overall user experience. First, while the model generally produces engaging and relevant responses to long user utterances, when the user utterance is short or contains ASR errors, the model digresses from task-oriented domains and fails to drive to the target domains. Second, we find that the model degenerates over time; the performance is usually good at the beginning of the interaction, while over time, the model produces less relevant and repetitive responses.
Hence, we terminate the social chat after the first two turns or when the user explicitly indicates their goal to alleviate these issues.

\subsection{DIY Task Retriever}

After constructing three queries described in \ref{ssec:diy_understanding} and obtaining the associated results, we check if the returned Wikihow articles are implied by the user utterance using an NLI-based reranker as mentioned above.
We take top-2 results of each query and re-rank all as our final displayed results. 

\subsection{Recipe Retriever}\label{subsec:recipe-retriever}
After tagging cooking-related entities, we run an \textbf{ordered} process of four steps to retrieve recipes via the provided Whole Food Market API. Specifically, when the previous one does not return any recipe, we move to the next step for continuing search.
\begin{compactenum}
    \item Using the tagged \textit{dish name} as dish name constraint.
    \item Using the tagged \textit{dish name} as cuisine constraint.
    \item Using all tagged \textit{cooking-related entities} as corresponding constraints.
    \item Using the last \emph{noun} recognized in the user utterance as the dish name constraint.
\end{compactenum}
The system replies no results only if all steps fail to retrieve any recipe.



\subsection{Task Content Responder}\label{subsubsec:task-responder}
The task content RG is designed to provide task instructions, ingredients, and other relevant content in the retrieved task or recipe. We use template-based methods to generate content. 
The user can navigate the content using a list of commands or touch screens. For example, the user can use ``show ingredients'' or ``go to the last step'' for navigation. We also add fun facts about recipe ingredients or tools to improve engagement. The fun facts are retrieved from a set of pre-defined fun facts crawled from the web using a simple TF-IDF model.

\subsection{Question Answering}\label{subsec:question-answering}
Users may ask various types of questions during task instructions.
After the question classifier identifies the user is asking a question, the question-answering (QA) RG focuses on finding the answer from the task content, recipe, or online resources. 
Four question-answering systems are employed: a rule-based system for ingredients quantity and substitution question in the recipe, two neural models for instructions-related questions in DIY and cooking domains, and an open-domain QA model for general questions supported by Amazon. For building task-specific neural QA models, we use BERT-based extractive QA model pre-trained on SQuAD \cite{rajpurkar2018know} and train our own conversational QA model. for recipe-related questions. 

\vspace{-3mm}
\paragraph{Model} 
To keep the conversational contexts in QA, we choose a conversational question answering model, where the input contains task context, previous conversation turns and user question. 
Two models are performed in our experiments: 1) BERT \cite{devlin-etal-2019-bert} and SciBert \cite{beltagy-etal-2019-scibert} for both conversational and traditional QA models.
They are all pre-trained on the CoQA dataset \cite{reddy-etal-2019-coqa} for conversational QA scenarios.
These models are further fine-tuned on our own dataset collected from logs or hand-written examples spanning multiple types of questions.

For traditional QA systems, the model input contains task context and questions, and the questions is previous conversation turns concatenated with user question. The symbol ``Q:'' and ``A:'' are added before each user utterance and each response respectively. 
Only last $k$ turns are selected as the input sequence due to the input length limit, where $k = \{1, 5\}$. 

\vspace{-3mm}
\paragraph{Data}
There are 4 question types in our training data: \emph{quantity-related}, \emph{time-related}, \emph{context-dependent} and \emph{others}. Quantity-related questions are about the ingredient quantity, such as ``{\it How many eggs to make the cake?}''; time-related questions are about cooking or task time, like ``{\it How long does it take to steam tomatoes?}''; context-dependent questions do not contain sufficient cues in the user question turn and can be answered depending on the conversation history. For example, in the user question turn ``{\it Where to place it?}'', the pronouns ``{\it it}'' must be referred from the previous dialogue.
Others indicate general questions that do not belong to the previous three categories.

\begin{table}[t!]
  \caption{Conversational QA performance. (Best in \textbf{bold}, second in \textit{italic}.)}
  \vspace{1mm}
  \centering
  \begin{tabular}{llccc}
    \toprule
    \bf Model & \bf \# History Turns & \bf  Fuzzy Match & \bf  F1 & \bf Exact Match (EM)\\
    \midrule
    BERT & No context &  \it 0.71 &  \it 0.65 & \it 0.46 \\
    BERT & Single previous turn &  \bf 0.73 &  \bf 0.66 &  \bf 0.48 \\
    BERT & 5 previous turns &  0.59 &  0.48 &  0.36 \\
    \midrule
    SciBERT & No context &  0.70 & 0.64 & 0.43 \\
    SciBERT & Single previous turn &  0.68 &  0.63 &  0.43 \\
    SciBERT & 5 previous turns &  0.58 &  0.50 &  0.42 \\
    \bottomrule
  \label{table:convQA-table}
  \end{tabular}
  \vspace{-3mm}
\end{table}

\begin{table}[t]
  \caption{F1 performance of two best results across different question types. (Best in {\bf bold}.)}
  \vspace{1mm}
  \centering
  \begin{tabular}{llccccc}
    \toprule
    \bf Model  & \bf \# History Turns & \bf Quantity &  \bf Time & \bf Context & \bf Others \\
    \midrule
    BERT & No context &  0.82 & \bf 0.88 & 0.46 &  0.35 \\
    BERT & Single previous turn & \bf 0.84 &  0.81 &  \bf 0.52 & \bf 0.39 \\
    \bottomrule
  \label{table:question_type_convQA-table}
  \end{tabular}
  \vspace{-3mm}
\end{table}

\vspace{-3mm}
\paragraph{Experiments} 

We perform two models in different history usage: no context, a single previous turn, and 5 previous turns, and evaluate the performance using three metrics: fuzzy match, F1 score (token-level match), and exact match (EM), where the fuzzy score \footnote{\url{https://github.com/seatgeek/thefuzz}} is the Levenstein distance between the gold answer and the predicted answer.
All performance is shown in Table~\ref{table:convQA-table}, where BERT without context and with a single previous turn (the first two rows) generally achieve better performance among all results.
To further analyze the performance of each question type, Table \ref{table:question_type_convQA-table} presents the detailed performance of these two methods.
It can be seen that BERT without context outperforms one with context only for time-related questions but performs worse for all other types.


\vspace{-3mm}
\paragraph{Discussion}
Although classical QA model can reasonably handle most questions, it cannot answer the questions with dependency. For the context-dependent questions, conversational BERT with single-turn history performs best.
Moreover, using a single turn is better than the 5-turn model, because incorporating too many history may introduce more noise.
Due to the conversational property, the BERT model with a single previous turn is selected and applied in the deployed system.

\section{Analysis}\label{sec:analysis}
This section aims to analyze the system performance by examining user ratings with various automatic metrics. Unless specified otherwise, the reported results are based on all user-rated conversations during the semi-finals period from February to March 2022. The data can help  understand how our bot performs and what we can do to improve the interactions in future competitions.

\subsection{Relationship between Rating and User Dialogue Intent}
\begin{figure}
     \centering
     \begin{subfigure}[b]{0.49\textwidth}
         \centering
          \includegraphics[width=\textwidth]{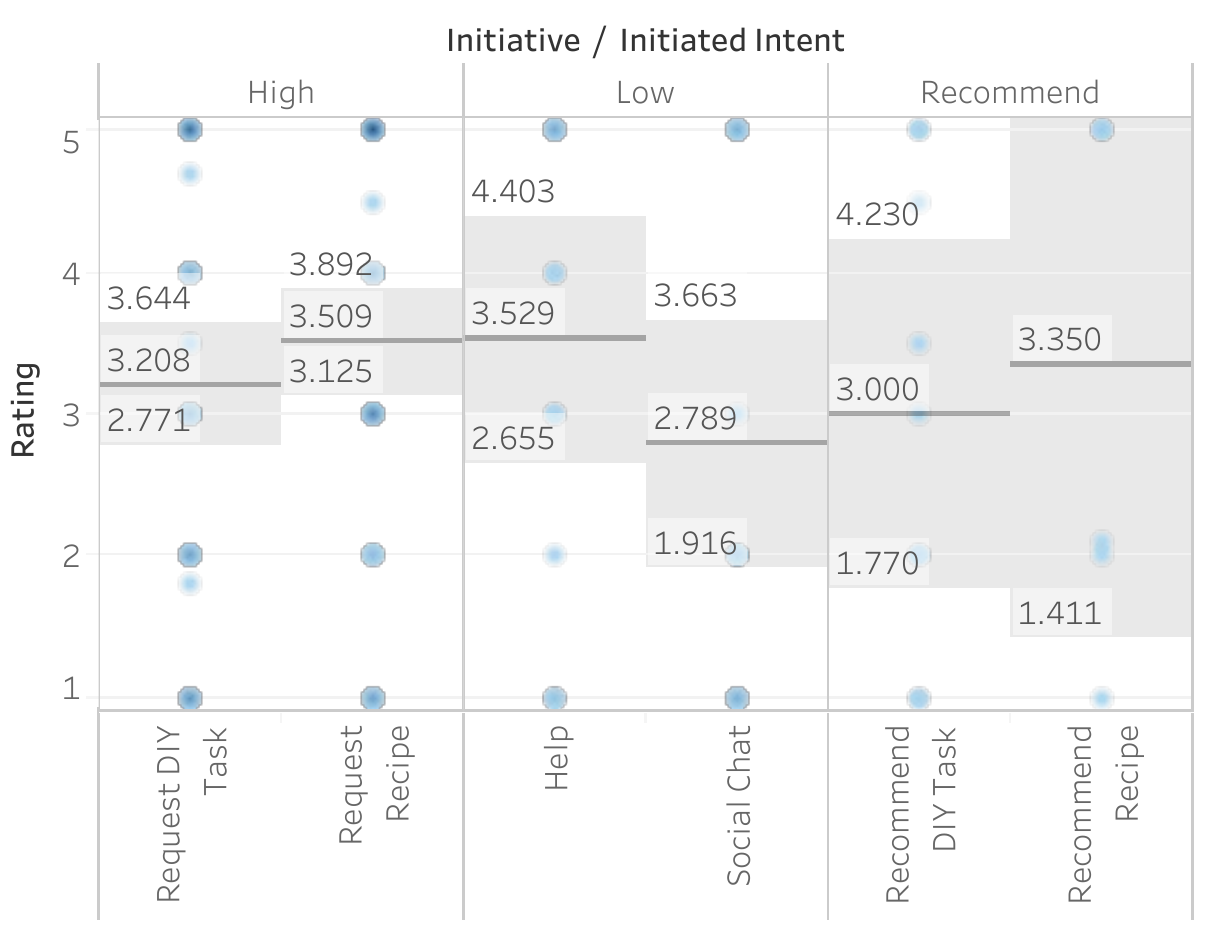}
          \caption{Density plot for initiated intent vs rating. Average rating with 95\% confidence interval.}
          \label{fig:intent-rating}
     \end{subfigure}
     \hfill
     \begin{subfigure}[b]{0.5\textwidth}
         \centering
          \includegraphics[width=\linewidth]{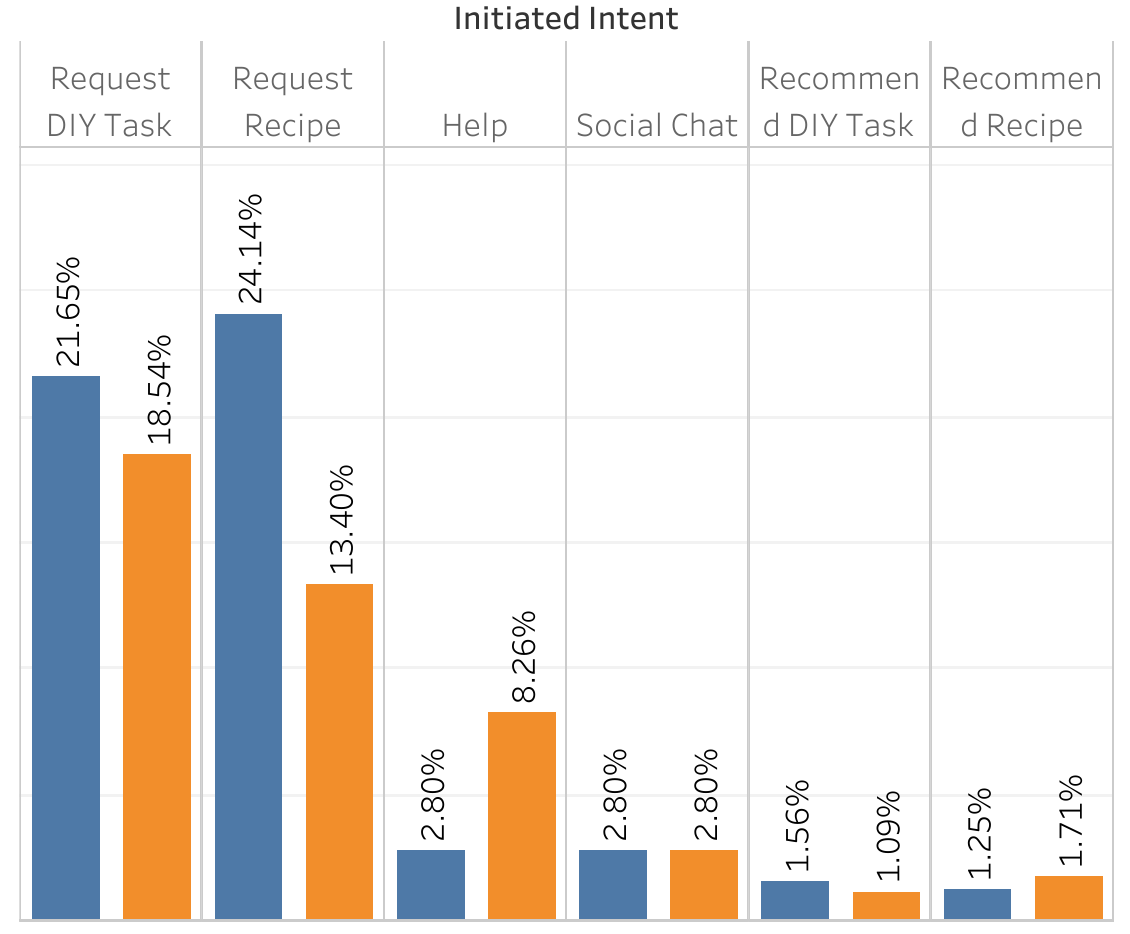}
          \caption{Conversation proportions with respect to  initial user intents for rated (blue) and unrated (orange) dialogues.} 
          \label{fig:intent_dist}
     \end{subfigure}
      \caption{Relationship between user ratings and initial intents.}
      \vspace{-4mm}
\end{figure}

To understand how initial user intents and the initiative level affects their ratings, we plot the relationship between user ratings and initiative across all initiated intents in Figure \ref{fig:intent-rating}.
The initiated intents are categorized into three initiative levels: \textit{high initiative}, \textit{low initiative}, and \textit{recommendation}, so that we can analyze the rating difference for different user behaviors. 


Figure \ref{fig:intent-rating} tells that users starting to \textit{request recipe} had a higher average rating than those who \textit{request DIY task} in the high-initiative part. 
The probable reason is that searching the recipe is easier than searching the right DIY task due to its limited scope, easy-to-follow instructions, and lower ambiguity in the user utterances.
\textit{Social chat} intents have the lowest average rating, likely because the users are not expecting to have a long chit-chat with the bot. 
The recommendation intents have a high variance in rating, probably due to diverse user expectations and preferences for recommendation.

Figure \ref{fig:intent_dist} shows the distribution of initiated intents for both user-rated and unrated conversations.
It can be found that there are more high-initiative users (\textit{request recipe} and \textit{request DIY task}) than low-initiative users (\textit{help}, \textit{social chat}, and \textit{recommendation}) in both rated and unrated conversations.
Also, in the rated conversations, more users start with \textit{request recipe} than \textit{request DIY task}.
Because the cooking domain is much easier to complete, users may be more likely to rate the bot.
The distribution suggests that most users have their requests in mind, aligning with the goal of our taskbot.



\subsection{Relationship between Ratings and Response Generators}
We also examine the relationship between user ratings and the last used response generator before the conversation ends.
In addition to all RG in \ref{sec:response-generator}, we investigate their sub-categories for the analysis.
\textit{Show Explicit Options} shows the list of available options returned by task retrievers to users.
\textit{Show Alternative Options (or No Options)} shows pre-defined alternative options or indicates no available options to users when retrievers fail. \textit{Confirm Step} confirms the selected option  when user utterances are ambiguous. 
\textit{Task Content (Ingredients)} generates the ingredient list for the recipe. \textit{Task Completed} informs the last-step instruction and indicates the task completion.
\textit{Changing Task} is a rule-based fallback response generator when the user wants to change a task. Note that users cannot change the task once the task is started based on Alexa Prize rules.

\begin{figure}[t!]
  \centering
  \includegraphics[width=.9\linewidth,keepaspectratio]{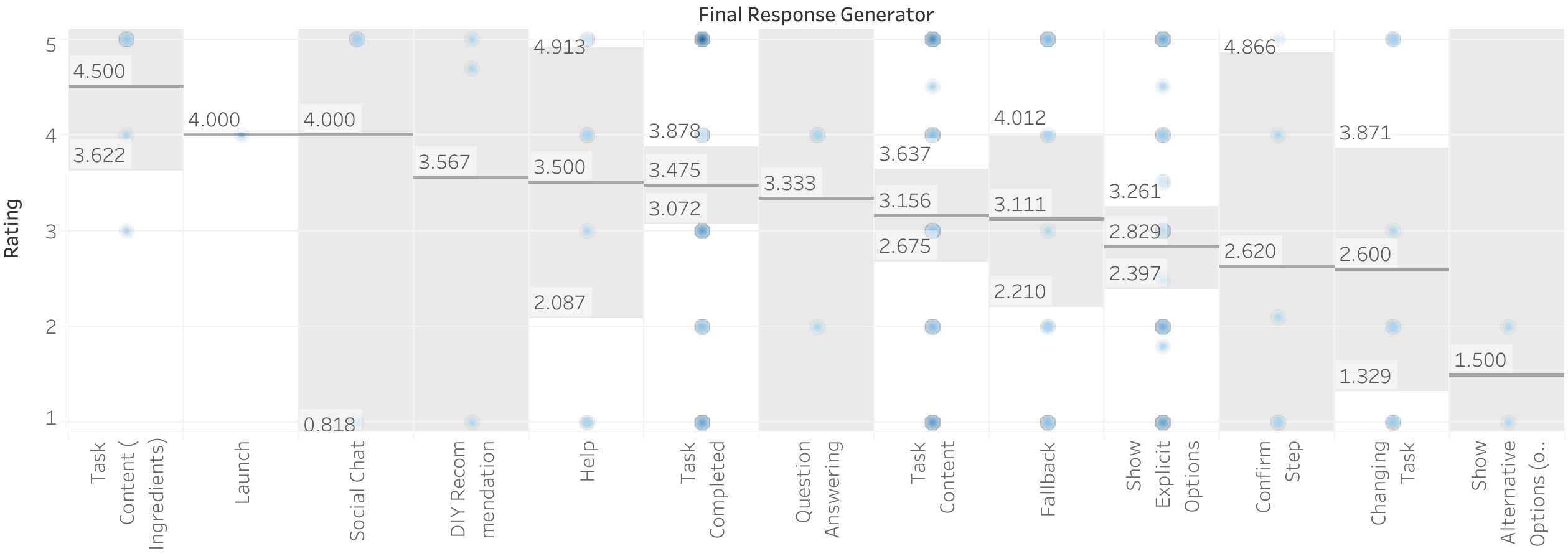}
  \vspace{-1mm}
  \caption{Density plot for final response generator vs rating. Average rating is represented by a horizontal gray line with 95\% confidence interval as shaded region.}
  \vspace{-3mm}
  \label{fig:rg-rating}
\end{figure}

\begin{figure}[t!]
  \centering
  \includegraphics[width=.9\linewidth,keepaspectratio]{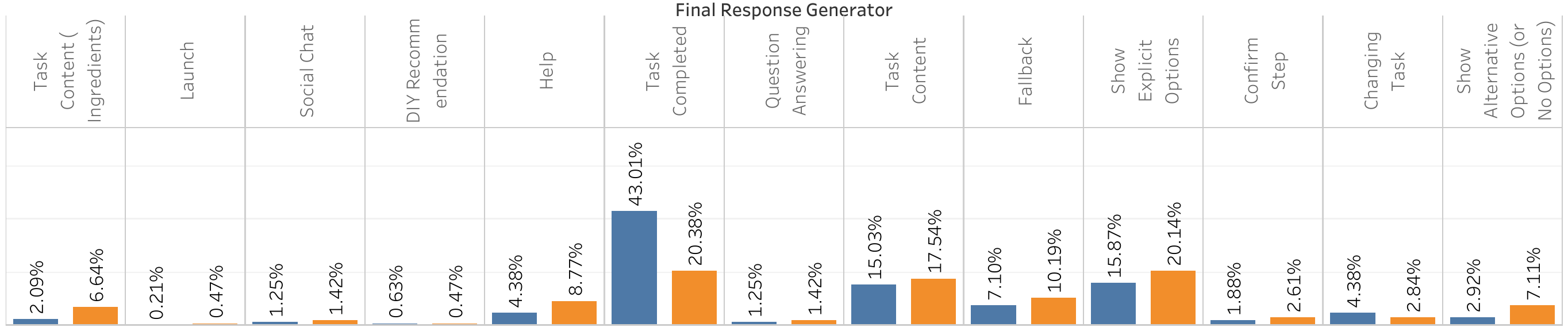}
  \caption{Conversation proportions in final generators for rated (blue) and unrated (orange) dialogues.}
  \label{fig:rg_dist}
  \vspace{-3mm}
\end{figure}

Figure \ref{fig:rg-rating} shows the relationship between user ratings and final response generators.
Overall, the users who start any task ({\it Task Content}) are more likely to give higher ratings.
We believe that the high rating for \textit{Task Completed} is due to user satisfaction with the task and their expectations being met.
Also, the high rating for \textit{Task Content (Ingredients)} may come from the informative content.
In contrast, it can be found that friction during the conversation would reduce user rating.
For example, \textit{Fallback}, \textit{Changing Task}, \textit{Show Alternative Options (or No Options)}, and \textit{Confirm Step} are used when the bot fails to understand user utterances or does not have enough information to complete the task.
\textit{Show Explicit Options} is polarized in ratings, probably because different user expectations. 
As expected, the unallowed {\it Changing Task} has a lower average rating.

Figure \ref{fig:rg_dist} shows the distribution of final response generators across both rated and unrated conversations.
More users can reach the \textit{Task Completed} stage, and they are willing to rate the bot. Following the observation that friction may reduce user rating, we find that users are less likely to rate the bot when they end with \textit{Help}, \textit{Fallback}, \textit{Show Alternative Options (or No Options)}, and \textit{Confirm Step}.


\section{Discussion and Future Work}\label{sec:discussion-and-future-work}

\vspace{-3mm}
\paragraph{Initiative}
Our taskbot aims to provide support for users who want to complete specific tasks, so we identify the user goal in the beginning to provide helpful content. However, we find that users may not always have clear goals in their minds, so asking users questions to elicit their goal may overwhelm users given the short time constraints of Alexa devices.
The future work in building systems should seamlessly transition from goal-seeking chatting to task-based dialogues.

\vspace{-3mm}
\paragraph{Conversational Content Presentation} 
Precisely and clearly presenting task instructions in a conversational manner is important for user engagement. However, we find that our current solutions suffer from two problems. First, template-based methods result in long, repetitive responses. Second, generative summarization models are prone to generating hallucination and even inappropriate content due to their neural nature. One potential future work is to build models that produce concise and coherent content in natural conversational styling.

\vspace{-3mm}
\paragraph{Error Recovery} 
Errors either in the user input (e.g., ASR errors,  wrong task, or recipe selection) or in the system (e.g., misidentified user intent, failures of task content retrieval) would result in deterioration of user experience. To mitigate this issue, we use fallback mechanisms that provide lists of possible commands. However, such list may not always be comprehensive. In addition, error recovery slows down the dialogue and hurts user engagement. Hence, sophisticated error recovery mechanisms based on dialogue context and user feedback are needed in the future work.

\subsubsection*{Acknowledgments}
Thanks to Shang-Yu Su for all of his guidance and support.
We would like to acknowledge the financial and technical support from Amazon.

\bibliographystyle{unsrt}
\bibliography{anthology, ref}

\begin{thebibliography}{10}

\bibitem{khatri2018advancing}
Chandra Khatri, Behnam Hedayatnia, Anu Venkatesh, Jeff Nunn, Yi~Pan, Qing Liu,
  Han Song, Anna Gottardi, Sanjeev Kwatra, Sanju Pancholi, et~al.
\newblock Advancing the state of the art in open domain dialog systems through
  the alexa prize.
\newblock {\em arXiv preprint arXiv:1812.10757}, 2018.

\bibitem{paranjape2020neural}
Ashwin Paranjape, Abigail See, Kathleen Kenealy, Haojun Li, Amelia Hardy, Peng
  Qi, Kaushik~Ram Sadagopan, Nguyet~Minh Phu, Dilara Soylu, and Christopher~D
  Manning.
\newblock Neural generation meets real people: Towards emotionally engaging
  mixed-initiative conversations.
\newblock {\em arXiv preprint arXiv:2008.12348}, 2020.

\bibitem{lewis-etal-2020-bart}
Mike Lewis, Yinhan Liu, Naman Goyal, Marjan Ghazvininejad, Abdelrahman Mohamed,
  Omer Levy, Veselin Stoyanov, and Luke Zettlemoyer.
\newblock {BART}: Denoising sequence-to-sequence pre-training for natural
  language generation, translation, and comprehension.
\newblock In {\em Proceedings of the 58th Annual Meeting of the Association for
  Computational Linguistics}, pages 7871--7880, Online, July 2020. Association
  for Computational Linguistics.

\bibitem{williams-etal-2018-broad}
Adina Williams, Nikita Nangia, and Samuel Bowman.
\newblock A broad-coverage challenge corpus for sentence understanding through
  inference.
\newblock In {\em Proceedings of the 2018 Conference of the North {A}merican
  Chapter of the Association for Computational Linguistics: Human Language
  Technologies, Volume 1 (Long Papers)}, pages 1112--1122, New Orleans,
  Louisiana, June 2018. Association for Computational Linguistics.

\bibitem{devlin-etal-2019-bert}
Jacob Devlin, Ming-Wei Chang, Kenton Lee, and Kristina Toutanova.
\newblock {BERT}: Pre-training of deep bidirectional transformers for language
  understanding.
\newblock In {\em Proceedings of the 2019 Conference of the North {A}merican
  Chapter of the Association for Computational Linguistics: Human Language
  Technologies, Volume 1 (Long and Short Papers)}, pages 4171--4186,
  Minneapolis, Minnesota, June 2019. Association for Computational Linguistics.

\bibitem{yu-yu-2021-midas}
Dian Yu and Zhou Yu.
\newblock {MIDAS}: A dialog act annotation scheme for open domain
  {H}uman{M}achine spoken conversations.
\newblock In {\em Proceedings of the 16th Conference of the European Chapter of
  the Association for Computational Linguistics: Main Volume}, pages
  1103--1120, Online, April 2021. Association for Computational Linguistics.

\bibitem{bosselut-etal-2019-comet}
Antoine Bosselut, Hannah Rashkin, Maarten Sap, Chaitanya Malaviya, Asli
  Celikyilmaz, and Yejin Choi.
\newblock {COMET}: Commonsense transformers for automatic knowledge graph
  construction.
\newblock In {\em Proceedings of the 57th Annual Meeting of the Association for
  Computational Linguistics}, pages 4762--4779, Florence, Italy, July 2019.
  Association for Computational Linguistics.

\bibitem{shi2019simple}
Peng Shi and Jimmy Lin.
\newblock Simple bert models for relation extraction and semantic role
  labeling.
\newblock {\em arXiv preprint arXiv:1904.05255}, 2019.

\bibitem{brown2020language}
Tom Brown, Benjamin Mann, Nick Ryder, Melanie Subbiah, Jared~D Kaplan, Prafulla
  Dhariwal, Arvind Neelakantan, Pranav Shyam, Girish Sastry, Amanda Askell,
  et~al.
\newblock Language models are few-shot learners.
\newblock {\em Advances in neural information processing systems},
  33:1877--1901, 2020.

\bibitem{roller2020recipes}
Stephen Roller, Emily Dinan, Naman Goyal, Da~Ju, Mary Williamson, Yinhan Liu,
  Jing Xu, Myle Ott, Kurt Shuster, Eric~M Smith, et~al.
\newblock Recipes for building an open-domain chatbot.
\newblock {\em arXiv preprint arXiv:2004.13637}, 2020.

\bibitem{rajpurkar2018know}
Pranav Rajpurkar, Robin Jia, and Percy Liang.
\newblock Know what you don't know: Unanswerable questions for squad.
\newblock {\em arXiv preprint arXiv:1806.03822}, 2018.

\bibitem{beltagy-etal-2019-scibert}
Iz~Beltagy, Kyle Lo, and Arman Cohan.
\newblock {S}ci{BERT}: A pretrained language model for scientific text.
\newblock In {\em Proceedings of the 2019 Conference on Empirical Methods in
  Natural Language Processing and the 9th International Joint Conference on
  Natural Language Processing (EMNLP-IJCNLP)}, pages 3615--3620, Hong Kong,
  China, November 2019. Association for Computational Linguistics.

\bibitem{reddy-etal-2019-coqa}
Siva Reddy, Danqi Chen, and Christopher~D. Manning.
\newblock {C}o{QA}: A conversational question answering challenge.
\newblock {\em Transactions of the Association for Computational Linguistics},
  7:249--266, 2019.

\end{thebibliography}
\end{document}